\documentclass[runningheads]{llncs}
\usepackage{amsmath}
\usepackage{marvosym}
\usepackage{hyperref}
\usepackage{array}
\usepackage[table]{xcolor}
\usepackage{makecell}
\usepackage{booktabs} 
\usepackage{multirow} 
\usepackage[T1]{fontenc}
\usepackage{graphicx,verbatim}
\begin{document}
\title{Can Agents Distinguish Visually Hard-to-Separate Diseases in a Zero-Shot Setting? A Pilot Study}
\titlerunning{A Pilot Study of Agents for Visually Confounded Diseases}
%
\author{Zihao Zhao\orcidID{0009-0001-8044-7683} \and
Frederik Hauke\orcidID{0000-0003-3434-5720} \and \\
Juliana De Castilhos \orcidID{0000-0001-9966-8657} \and \\
Sven Nebelung \orcidID{0000-0002-5267-9962} \and
Daniel Truhn $^{(\textrm{\Letter})}$\orcidID{0000-0002-9605-0728}
}
\authorrunning{Z. Zhao et al.}
%
\institute{
Department of Diagnostic and Interventional Radiology, University Hospital Aachen, 52074 Aachen, Germany \\
\email{dtruhn@ukaachen.de}}

\maketitle              
\begin{abstract}
The rapid progress of multimodal large language models (MLLMs) has led to increasing interest in agent-based systems. While most prior work in medical imaging concentrates on automating routine clinical workflows, we study an underexplored yet clinically significant setting: distinguishing visually hard-to-separate diseases in a zero-shot setting. 
We benchmark representative agents on two imaging-only proxy diagnostic tasks, (1) melanoma vs.\ atypical nevus and (2) pulmonary edema vs.\ pneumonia, where visual features are highly confounded despite substantial differences in clinical management.
We introduce a multi-agent framework based on contrastive adjudication.
Experimental results show improved diagnostic performance (an 11-percentage-point gain in accuracy on dermoscopy data) and reduced unsupported claims on qualitative samples, although overall performance remains insufficient for clinical deployment.
We acknowledge the inherent uncertainty in human annotations, conservative binary setting, and the absence of clinical context, which further limit the translation to real-world settings. 
Within this controlled setting, this pilot study provides preliminary insights into zero-shot agent performance in visually confounded scenarios. Our code is available at \href{https://github.com/TruhnLab/Contrastive-Agent-Reasoning}{https://github.com/TruhnLab/Contrastive-Agent-Reasoning}.

\end{abstract}

\section{Introduction}
Agents are systems that receive observations and perform goal-directed actions (e.g., making decisions or plans) to optimize specific objectives~\cite{akerkar2026ai}. In recent years, multimodal large language models (MLLMs)~\cite{chen2024internvl,bai2025qwen3,team2025gemma,vteam2025glm45vglm41vthinkingversatilemultimodal,gemini3_software} have advanced rapidly in capability, making them well-suited to serve as such agents. 
Hence, a growing number of studies~\cite{wang2025survey} have adopted MLLMs as agents and integrated them into clinical practice, aiming to automate routine workflows or assist physicians in decision-making. However, most existing systems~\cite{wang2025survey} assume always-on human-agent collaboration, while some recent studies~\cite{vaccaro2024combinations,rosbach2025automation} have questioned this assumption and suggested selective collaboration.
\begin{figure}
    \centering
    \includegraphics[width=0.95\textwidth]{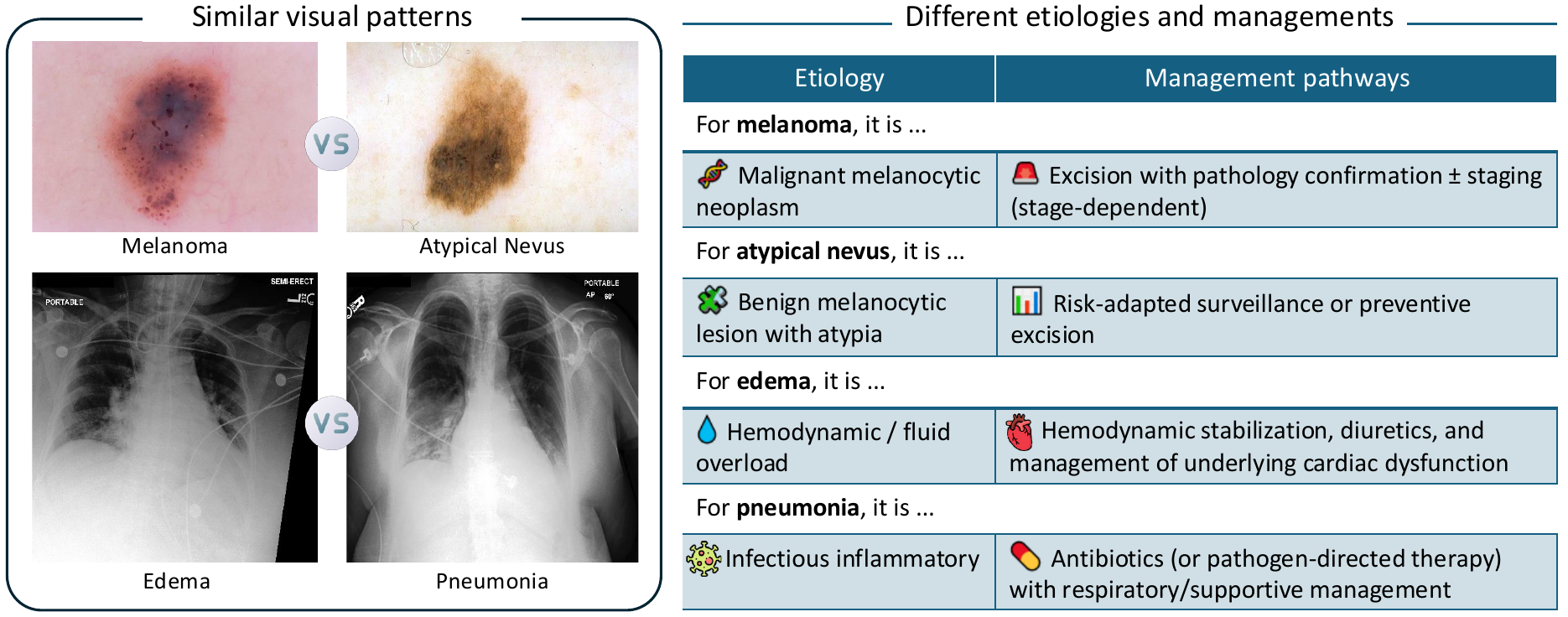}
    \caption{
    \textbf{Illustration of hard-to-separate disease pairs.}
Despite highly overlapping visual patterns, their typical etiologies and managements differ significantly, which makes imaging-only differentiation challenging and high-stakes.
}
    \label{fig:teaser}
\end{figure}

Motivated by this, we focus on a less explored yet clinically critical setting: distinguishing visually hard-to-separate diseases with distinct etiologies and management pathways. Fig.~\ref{fig:teaser} provides a straightforward illustration. 
In dermoscopy, distinguishing melanoma from atypical nevus is a well-recognized diagnostic challenge~\cite{jitian2021clinical} because both are melanocytic lesions and can share asymmetry and irregular borders. 
As a fatal cancer, concern about melanoma is a common reason for dermatology consultation and skin checks~\cite{mahama2024lived}.
Likewise, on chest radiographs, edema and pneumonia are frequently confused because both present with lung opacities and diffuse haziness~\cite{gluecker1999clinical}.
This makes separation challenging for inexperienced radiologists. Despite visual similarity, their management methods are different.
This discrepancy naturally raises a critical question: can current MLLM-based agents distinguish such visually confounded diseases in a zero-shot manner, and thus assist junior clinicians?

However, current MLLM-based agents are still far from satisfactory. 
Under high-ambiguity scenarios, a single agent could prematurely favor one hypothesis, and produce overconfident statements to support this hypothesis~\cite{huang2025survey}.
Recent efforts to alleviate this issue rely on additional annotated data for fine-tuning~\cite{leng2024mitigating,liang2025anatomical}, repeated sampling to assess response consistency~\cite{wienholt2025hallucination,zhang2025radflag}, agent collaboration with tool-use mechanisms~\cite{feng2026pass}, or training recipes designed to mitigate overconfidence~\cite{jiang2025knowing}. Such approaches, however, either introduce external tool dependencies or require additional training, and are not well-aligned with our zero-shot setting. 
To this end, we propose a novel multi-agent system named \textbf{C}ontrastive \textbf{A}gent \textbf{RE}asoning (CARE).   
The underlying philosophy is straightforward: even experienced human experts reason by contrast~\cite{buccinca2025contrastive}. Specifically, explaining why the case supports one side and why it opposes the other. In CARE, one agent argues for melanoma while another argues for atypical nevus, and a third agent acts as a judge and summarizes their contrastive arguments to make the final diagnosis. 

To summarize, our main contributions are listed as follows:
\begin{itemize}
    \item To the best of our knowledge, this is among the first studies to benchmark MLLM-based agents on visually confounded diseases in a zero-shot setting.
    \item We propose CARE, a novel multi-agent system that improves agent performance by explicitly structuring disagreement, without additional training.
    \item Extensive benchmarking across two imaging modalities demonstrates statistically significant improvements over baselines (p < 0.0001 for dermoscopy; p < 0.001 for chest X-rays, McNemar and permutation tests), yet overall performance remains below the requirements for clinical use.

\end{itemize}

\section{Method}

We first formulate the problem setting in Section~\ref{sec:problem_state}. Then, we detail the principle of CARE (Section~\ref{sec:CARE_detail}). A brief analysis is provided in Section~\ref{sec:math_explan}.

\subsection{Problem Formulation}\label{sec:problem_state}
In this study, we focus on image-only prediction of report-derived labels under a forced exclusive-or (XOR) setting.
Let \( x \in \mathcal{X} \) denote a medical image (e.g., dermoscopy or chest radiograph), and let \( y \in \{A, B\} \) denote the corresponding diagnostic label extracted from a human-written report. The XOR criterion enforces mutual exclusivity between the two labels, thereby formulating the task as a controlled binary classification problem. Although this setup simplifies evaluation, it does not reflect the potential co-occurrence of conditions (e.g., edema and pneumonia) in real clinical practice.
Our setting is intentionally constrained to zero-shot inference: no task-specific fine-tuning and no annotated supervision. The goal is to output a diagnostic decision, together with visual evidences and an interpretable reasoning path. This setting differs from standard image classification, as a single probability score only provides limited decision support.

\begin{figure}[tbp]
    \centering
    \includegraphics[width=0.93\textwidth]{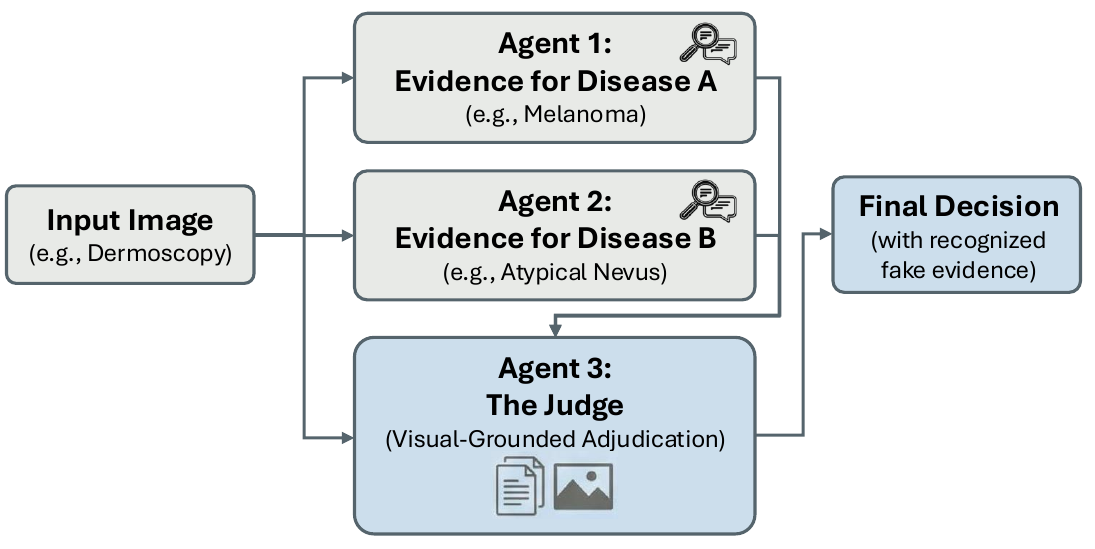}
    \caption{The overview of \textbf{Contrastive Agent Reasoning (CARE).}
Two disease-specific agents generate opposing evidence from the same input image (e.g., melanoma vs. atypical nevus). A judge agent adjudicates the arguments, flags unsupported evidence, and outputs the final diagnosis in a training-free, zero-shot setting.}
    \label{fig:method}
\end{figure}

\subsection{Contrastive Multi-Agent Reasoning}\label{sec:CARE_detail}
As illustrated in Fig.~\ref{fig:method}, 
CARE is training-free and does not require external tools or fine-tuning, but uses a structured multi-call inference with three roles.
Two disease-specific agents independently generate evidence conditioned on opposing diagnostic hypothesis, while a third agent adjudicates between their arguments.

\textbf{Role-conditioned evidence generation.}
Each disease-specific agent operates under a strict role condition. It must interpret the image solely from the perspective of its assigned diagnosis and enumerate visual evidence supporting that hypothesis. Agents are prohibited from making a final diagnosis. This role conditioning ensures that each agent is dedicated solely to evidence generation, rather than decision making.
As a result of this design choice, unsupported evidence becomes hypothesis-consistent but potentially image-inconsistent. Evidence generated for disease \(A\) could contradict or undermine disease \(B\), and vice versa. These inconsistencies enable contrastive adjudication.

\textbf{Visual-grounded judgment.}
The judge agent receives three inputs: the original medical image \( x \), the evidence set \( E_A \) generated under hypothesis \( y = A \), and the evidence set \( E_B \) generated under hypothesis \( y = B \). Its task is to evaluate the plausibility of each evidence set by grounding the claims in the image.
The judge agent performs three functions: (i) cross-checking evidence against the image, (ii) identifying unsupported or contradictory claims, and (iii) weighing the remaining contrastive arguments to reach a final diagnosis. Importantly, the judge does not introduce new medical evidence. It only evaluates existing claims.

\subsection{Why Contrastive Reasoning Benefits}\label{sec:math_explan}
The rationale of contrastive reasoning can be illustrated probabilistically.
A single agent implicitly commits to a decision
\[
\hat{y} = \arg\max_{y \in \{A,B\}} p(y \mid x),
\]
which becomes unstable when \( p(A \mid x) \approx p(B \mid x) \), as is common in visually confounded cases.
Hallucinated evidence, while often self-consistent within a single hypothesis, tends to become inconsistent across mutually exclusive hypotheses. Hence, instead of committing to a single hypothesis, CARE explicitly generates hypothesis-conditioned explanations:
\[
E_A \leftarrow \text{Agent}_1(x), \qquad
E_B \leftarrow \text{Agent}_2(x).
\]
A judge agent then evaluates the visual consistency of these competing explanations.
Let \( \mathcal{S}(x, E) \) denote an implicit visual-consistency assessment performed by the judge agent.
Conceptually, the final decision can be expressed as
\[
\hat{y}
= \arg\max_{y \in \{A,B\}}
\Big( \mathcal{S}(x, E_y) - \mathcal{S}(x, E_{\neg y}) \Big).
\]
Rather than relying on repetitive self-check~\cite{wienholt2025hallucination} or stochastic sampling~\cite{zhang2025radflag}, CARE induces contrast through structured role assignment.
By forcing explicit contrast between competing explanations, CARE distributes reasoning across competing hypotheses, mitigating premature commitment under visual ambiguity.
Note that CARE is implemented via structured prompting only. It does not compute explicit likelihoods or numerical functions in practice.

\section{Experimental Results}
\subsection{Dataset Curation and Implementations}
In this section, we curate two binary classification datasets from existing public datasets based on the following pipeline.

For \textbf{Melanoma vs. Atypical Nevus}, we curate dermoscopy images from the derm7pt dataset~\cite{kawahara2018seven}. We first enforce an XOR criterion, retaining only studies labeled as either atypical nevus or melanoma. We then exclude congenital and recurrent nevi, as these subtypes are often very difficult without clinical context. To mitigate class imbalance, we subsample atypical nevi, resulting in a final set of 509 studies with 257 atypical nevi and 252 melanomas.
For
\textbf{Edema vs. Pneumonia}, we curate chest X-rays from the MIMIC-CXR dataset~\cite{johnson2019mimic}. After the XOR criterion, we obtain 2,907 candidate studies with exclusive edema or pneumonia labels. Since report-derived labels can be noisy, we further exclude studies with low-confidence diagnostic statements by identifying expressions like ``can not be excluded'' based on associated radiology reports, yielding a final set of 1,739 studies (878 edema and 861 pneumonia).

\textbf{Implementations.} All curated data are used exclusively for evaluation (test-only) in this study. We provide no external tools for MLLM-based agents to fairly compare their capabilities. Experiments with open-source MLLMs are conducted on a single NVIDIA H100 GPU (96\,GB). Due to data privacy requirements, Gemini, instead of GPT, is selected as the default closed-source MLLM in our study. We obtain API access to Gemini models via the Vertex AI platform, as Physionet \href{https://physionet.org/news/post/gpt-responsible-use/}{requested}.
Accuracy (ACC), F1-score (F1), and Youden-Index (Youden)~\cite{youden1950index}, defined as $\text{Sensitivity} + \text{Specificity} - 1$, are reported. CARE is built upon Gemini-3-Flash by default in this study. For Gemini-3-Flash vs. CARE and Gemini-3-Pro vs. CARE, statistical significance is evaluated, using McNemar~\cite{mcnemar1947note} (for ACC) and permutation tests~\cite{collingridge2013primer} (for F1 and Youden).
\begin{table}[t]
\centering
\caption{Benchmarking results on visually confounded diagnostic tasks. 
\colorbox{gray!20}{Gray}: best single-agent method. \textbf{Bold}: best performance across all methods.
}
\label{tab:results}
\begin{tabular}{lcccccc}
\toprule
\multirow{2}{*}{\textbf{Method}} 
& \multicolumn{3}{c}{\makecell{\textbf{Melanoma vs.}\\\textbf{Atypical Nevus}}}
& \multicolumn{3}{c}{\makecell{\textbf{Edema vs.}\\\textbf{Pneumonia}}} \\
\cmidrule(lr){2-4} \cmidrule(lr){5-7}
& ACC & F1 & Youden & ACC & F1 & Youden \\

\midrule
SigLIP2~\cite{tschannen2025siglip}                             & 0.611 & 0.606 & 0.219 & 0.502 & 0.349 & 0.006 \\
BiomedCLIP~\cite{zhang2025multimodal}                               & 0.607 & 0.569 & 0.209 & 0.557 & 0.554 & 0.114 \\
\midrule
MedVLM-R1~\cite{pan2025medvlm}                           & 0.505 & 0.368 & 0.001 & 0.498 & 0.336 & -0.003 \\
InternVL3-14B~\cite{chen2024internvl}                       & 0.549 & 0.534 & 0.071 & 0.502 & 0.361 & 0.002 \\
Gemma-3-12B~\cite{team2025gemma}                         & 0.499 & 0.427 & 0.091 & 0.476 & 0.455 & -0.041 \\
Qwen3-VL-8B~\cite{bai2025qwen3}                         & 0.661 & 0.656 & 0.365 & 0.559 & 0.542 & 0.119 \\
Qwen3-VL-32B~\cite{bai2025qwen3}                        & 0.698 & 0.670 & 0.347 & 0.564 & 0.504 & 0.116 \\
GLM-4.6V-Flash~\cite{vteam2025glm45vglm41vthinkingversatilemultimodal}                  & 0.718 & 0.714 & 0.426 & 0.552 & 0.513 & 0.088 \\
\midrule
Gemini-3-Flash~\cite{gemini3_software}                      &  & & & &  &  \\
\quad Baseline                     & 0.665 & 0.630 & 0.328 & 0.602 & 0.539 & 0.197 \\
\quad Self-Check$^\dagger$ (2×) & 0.701 & 0.681 & 0.400 & 0.606 & 0.546 & 0.200 \\
\quad Self-Check$^\dagger$ (3×) & 0.719 & 0.704 & 0.436 & 0.610 & 0.551 & 0.208 \\
\quad Majority-Vote$^\ddagger$ (3×) & 0.686 & 0.660 & 0.376 & 0.601 & 0.533 & 0.195 \\
\quad Blind-CARE{*} & 0.739 & 0.729 & 0.476 & 0.618 & 0.563 & 0.228 \\
\quad CARE                          & \textbf{0.776} & \textbf{0.769} & \textbf{0.552} & 0.646 & 0.619 & 0.287 \\
\midrule
\rowcolor{gray!20}
Gemini-3-Pro~\cite{gemini3_software}                         & 0.741 & 0.734 & 0.482 & \textbf{0.709} & \textbf{0.709} & \textbf{0.419} \\
\bottomrule
\multicolumn{7}{l}{\makecell[c]{\scriptsize $^\dagger$ Multi-pass self-revision; \scriptsize $^\ddagger$ Independent sampling with majority aggregation;\\\ \scriptsize {*} Judge agent has no access to the original image.}}
\end{tabular}\label{tab:main_results}
\end{table}

\subsection{Quantitative Analysis}\label{sec:main_results}

Table~\ref{tab:results} presents a comprehensive comparison across three categories of models: CLIP-based vision-language models, open-source, and closed-source MLLMs. 

\textbf{Overall performance reveals substantial challenges.} CLIP-based models (SigLIP2 and BiomedCLIP) achieve modest performance on both tasks, indicating that simple vision-language alignment without reasoning capabilities is insufficient for these visually confounded diagnostic scenarios. Other single-agent systems show better performance but remain largely unsatisfactory, with most of them achieving only 50-70\% accuracy. This highlights the difficulty of distinguishing visually confounded diseases.

\textbf{CARE yields consistent improvements.} Built upon Gemini-3-Flash, CARE framework demonstrates notable performance gains. On the melanoma vs. atypical nevus task, CARE achieves 77.6\% accuracy with a Youden Index of 0.552, representing an improvement of over 11 percentage points compared to the single-agent Gemini-3-Flash baseline (66.5\% accuracy, 0.328 Youden Index). Notably, the difference between CARE and  Gemini-3-Pro is not statistically significant on this dataset, based on McNemar (p-value = 0.192) and permutation (p-value = 0.192 for both F1 and Youden) tests. 
On the edema vs. pneumonia task, while CARE still outperforms its base model (64.6\% vs. 60.2\%), it falls short of Gemini-3-Pro's 70.9 \%.
Nevertheless, the improvement over Gemini-3-Flash is still significant, with p-value < 0.001 on all three metrics.
\begin{figure}[t]
    \centering
    \includegraphics[width=0.95\textwidth]{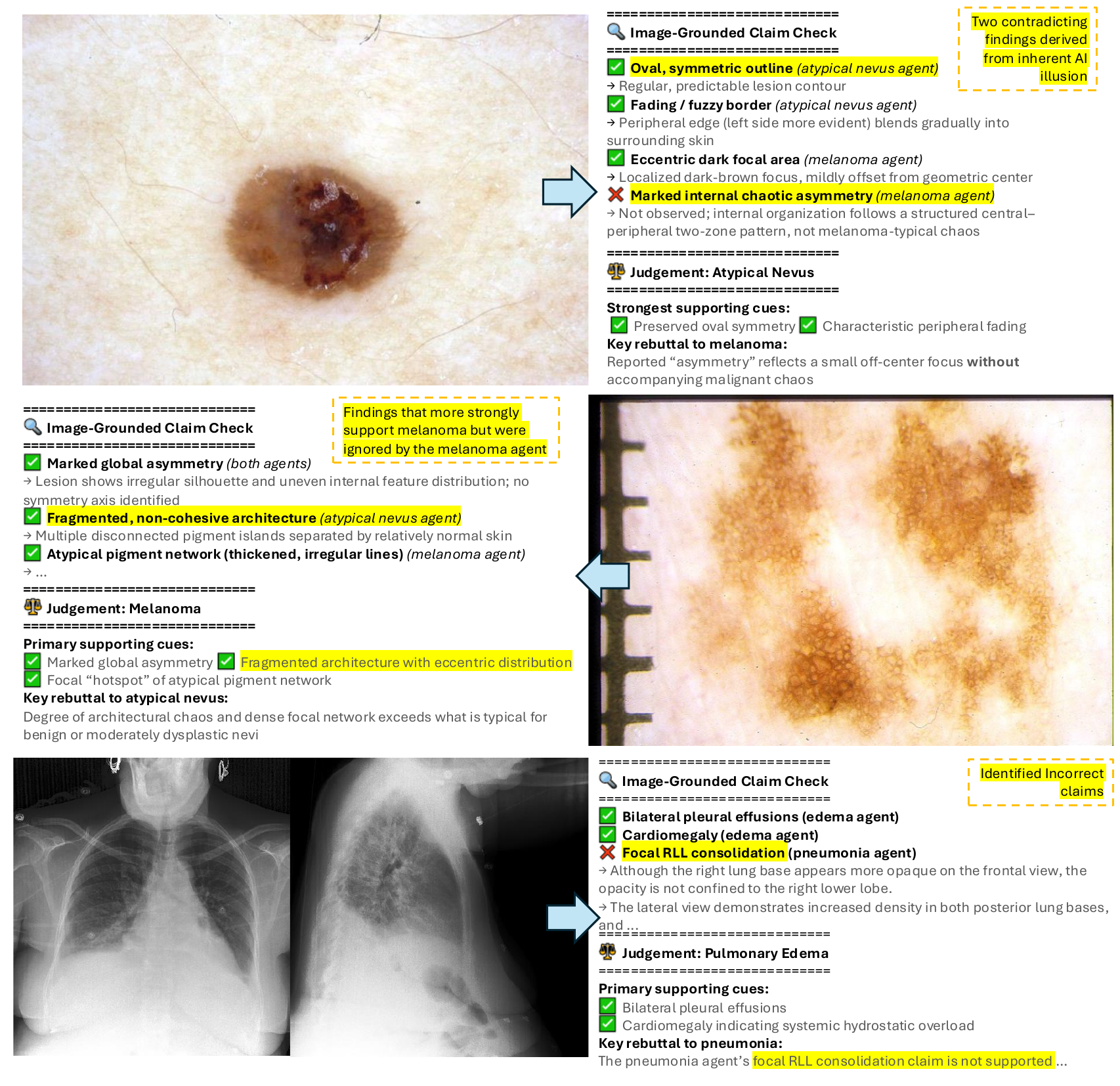}
    \caption{
Representative cases showing how CARE exposes contradictory findings, implements cross-agent evidence recalibration, and verifies claims against the image.
}
    \label{fig:judge_strengths}
\end{figure}

\textbf{Ablation studies validate the design of CARE.} We design three types of ablation variants to justify the contribution of our proposed CARE. 
Self-Check (2×) and Self-Check (3×) prompt Gemini-3-Flash to perform multi-pass self-reflection before producing a final decision, with two and three sequential reasoning passes, respectively. 
Majority-Vote (3×) independently queries Gemini-3-Flash three times and aggregates the final prediction via majority voting.
These two types of variants show only limited improvement over the single-call baseline, particularly on chest radiographs. Notably, Self-Check (3×) and Majority-Vote (3×) are compute-matched to CARE, as each method invokes the API three times per case. Their marginal gains indicate that performance improvement is not merely attributable to increased sampling or ensembling, but rather to the structured contrastive reasoning mechanism introduced by CARE. 
Interestingly, Majority-Vote tends to perform worse than Self-Check. One possible explanation is that errors in visually confounded cases may be correlated rather than purely random. If similar interpretations are repeatedly produced, simple aggregation may provide a limited corrective effect. 
We further evaluate the third type of variant, Blind-CARE, in which the judge agent has access only to the textual arguments from the two specialist agents. Blind-CARE achieves 73.9\% accuracy on melanoma, outperforming the two Self-Check variants but remaining inferior to CARE. This suggests that direct access to visual evidence is essential for detecting fabricated or misinterpreted claims and for effective adjudication. 

\subsection{Qualitative Analysis}

This section illustrates how CARE behaves in representative cases via Fig.~\ref{fig:judge_strengths}. First, CARE can detect contradictory findings. In the first case, the melanoma agent claims ``marked internal chaotic asymmetry'' that conflicts with the lesion’s overall symmetric architecture, which the judge explicitly flags and rejects via image-grounded claim checking, yielding the correct atypical nevus decision. 
Second, CARE enables \textbf{cross-agent evidence recalibration}. In visually confounded cases, certain findings may be compatible with both hypotheses. However, their specific morphological characteristics and spatial distribution determine which diagnosis they more strongly support. In our example, a fragmented and non-cohesive architecture is mistakenly interpreted during evidence generation as favoring atypical nevus. The judge re-evaluates this finding and correctly recalibrates its diagnostic weight toward melanoma.
Finally, CARE can identify an unsupported localized claim from the pneumonia agent by cross-checking against multi-view evidence, preventing a false pneumonia prediction and correctly concluding pulmonary edema.
Overall, CARE reduces ungrounded claims in the presented examples.

\section{Discussion and Conclusion}
In this study, we benchmark multiple MLLM-based agents and find their performance limited. Some methods even achieve a negative Youden index, indicating performance below chance level. This study has several limitations. First, label quality is imperfect. While part of the dermoscopy data was histologically verified~\cite{kawahara2018seven}, the rest relied on expert diagnosis. 
Chest X-ray labels were automatically extracted from radiology reports and therefore reflect report impressions rather than an independent reference standard (e.g., CT or clinical adjudication), introducing evaluation uncertainty.
Our mutually exclusive setting also conflicts with the fact that patients may have both edema and pneumonia. 
Second, agents were evaluated without external tools. Future work could investigate whether segmentation models or image retrieval could help improve performance.

In conclusion, we investigate a clinically important yet underexplored topic through imaging-only proxy tasks. 
We propose CARE, a multi-agent system improving zero-shot diagnostic performance in a training-free setting.
Findings suggest that structuring disagreement and image-based verification are essential for performance gain, offering insights for future multi-agent system design.
Nevertheless, our overall benchmarking results underscore that current MLLM-based agents are not yet suitable for clinical translation, highlighting the need for further methodological advances and more rigorous evaluation.

\begin{credits}
\subsubsection{\ackname} This research project is funded by the European Union Research and Innovation Programme (Horizon Europe - European Research Council, SAGMA – GA 101222556). The authors gratefully acknowledge the computing time provided to them at the NHR Center NHR4CES at RWTH Aachen University (project number p0021834). This is funded by the Federal Ministry of Research, Technology and Space , and the state governments participating on the basis of the resolutions of the GWK for national high performance computing at universities (www.nhr-verein.de/unsere-partner). The data used in this publication was managed using the research data management platform Coscine (http://doi.org/10.17616/R31NJNJZ) with storage space of the Research Data Storage (RDS) (DFG: INST222/1261-1) and DataStorage.nrw (DFG: INST222/1530-1) granted by the DFG and Ministry of Culture and Science of the State of North Rhine-Westphalia.

\subsubsection{\discintname}
Daniel Truhn holds shares in StratifAI and Synagen. He has received honoraria from Bayer, AstraZeneca, Philips, Roche, Pfizer, and Gilead. 
All other authors declare no conflicts of interest.
\end{credits}

%
%
%
%
%
%
\bibliographystyle{splncs04}
\bibliography{ref}
\end{document}